\documentclass{article}

\usepackage[preprint]{tmlr}

\usepackage{graphicx}
\usepackage{booktabs}
\usepackage{amsmath}
\usepackage{amssymb}
\usepackage{xcolor}
\usepackage{microtype}
\usepackage{tikz}
\usetikzlibrary{shapes.geometric, arrows.meta, positioning, calc}

\usepackage[hidelinks]{hyperref}
\usepackage{url}

\graphicspath{{figures/}}

\title{Knowing in Advance When an Evolutionary Outer Loop Will Not Help:\\
A Pre-Registered Cheap-Baseline Screening Rule}

\author{\name Ramchand Kumaresan \\
      \addr Murai Labs}

\begin{document}

\maketitle

\begin{abstract}
We introduce a pre-registered screening rule that decides, \emph{before any
implementation}, whether an evolutionary / population / lifecycle outer loop over
neural-network parameters or structure is worth building. Such outer loops cost
$10^2$--$10^3\times$ their gradient inner loop, yet whether they beat a cheap
single-shot alternative is usually discovered only after the expense is paid. Our
rule computes, at a Phase-0 gate, a single number---the recovery
$R = s/G$, the best single-shot gradient/curvature statistic's gain $s$ divided by
the best gain $G$ of any cheap method evaluated---and prescribes skipping the outer
loop when $R \geq 90\%$. We validate the rule on a within-lab series of
pre-registered outer-loop bets (two analyzed cases plus a disclosed file drawer):
in both analyzed cases a static or single-shot computation captured the effect on
the project's own metric, the gate fired ($R \approx 1.0$ in both cases;
${\approx}\,0.95$ under a stricter metric on one), and the outer loop was
abandoned---including one case where a companion factorial decomposition
localizes the apparent win to a static substrate change with the evolutionary
lifecycle contributing no detectable gain. On one project the gate cost
${\sim}50$--$70$ GPU-hours and screened out an
estimated ${\sim}400{+}$ GPU-hours (first cell only) plus weeks of
implementation---a ${\sim}6$--$8\times$ saving. The rule is prospectively
falsifiable: a task with $R < 90\%$ where the outer loop still fails to beat
single-shot would refute it.
\end{abstract}

\section{Introduction}

Evolutionary, population-based, and lifecycle \emph{outer loops} that search over
network parameters, masks, routes, or structure are an attractive idea: they can
optimize objectives a gradient inner loop cannot directly serve. They are also
expensive---commonly $10^2$--$10^3\times$ the cost of the inner loop they wrap. The
practical question a lab faces is therefore not ``\emph{can} an outer loop help?''
but ``\textbf{how do you know, cheaply and in advance, whether it will?}''

This paper answers that question with a pre-registered discipline and reports the
outcome of applying it to our own work. The contribution is \textbf{methodological},
not a new algorithm; the individual negative results the discipline produced are, on
their own, unsurprising (see Section~\ref{sec:related}). What is new is the packaged,
evidenced, prospectively-falsifiable \emph{decision procedure}.

\paragraph{Contributions.}
\begin{itemize}
\item \textbf{A Cheap-Baseline-Falsification gate (Section~\ref{sec:rule}):} a
  pre-registered Phase-0 protocol that runs the cheapest plausible non-mechanism
  explanations (random, single-shot heuristic, static/non-evolutionary) on the
  \emph{same metric the paper will use}, before the outer loop is built.
\item \textbf{An operational screening rule (Section~\ref{sec:rule}):} skip the
  outer loop when recovery $R = s/G \geq 90\%$, where $s$ is the best single-shot
  gradient/curvature statistic's gain and $G$ is the best gain of any cheap method
  evaluated. The $90\%$ threshold is a conservative convention placed below the
  observed kills, and the rule is prospectively falsifiable.
\item \textbf{Validation on a within-lab series (Sections~\ref{sec:caseone}--\ref{sec:casetwo}):}
  two analyzed cases, alongside a disclosed file drawer. In both analyzed cases the
  gate screened out the outer loop, including a case whose submitted companion
  factorial isolates the outer loop as the non-contributor. (We do not claim
  independence---Section~\ref{sec:limitations}.)
\item \textbf{A GPU-hours-saved ledger (Section~\ref{sec:ledger})} quantifying what
  the gate buys.
\end{itemize}

We are explicit about what this paper is \emph{not}: it is not a claim that
evolutionary outer loops never help. Section~\ref{sec:positive} states the
converse---the condition under which they \emph{should} help---as a hypothesis with
a named, pre-registered successor design whose first screen is now reported; no
claim here depends on its unrun branches.

\section{Related Work and Positioning}
\label{sec:related}

We position against three lines of work; none is contested by this paper, which is
why our individual negatives are unsurprising and our contribution is the screening
procedure rather than any one result.

\paragraph{Evolution strategies on smooth objectives.} ES is a competitive
black-box optimizer when gradients are unavailable or the landscape is rugged
\citep{salimans2017es, such2017deep, lehman2018surprising}. On smooth,
differentiable objectives the inner optimizer already serves, ES is known to be at
best competitive and often worse: \citet{zhang2017esrelationship} show ES estimates
the gradient of a Gaussian-smoothed objective, so when the per-minibatch gradient is
already accessible its overhead is unwarranted, and \citet{lenc2019nondiff} find ES
competes with gradient-based search only on small problems. Our cases sit in this
regime, and the screening rule operationalizes \emph{detecting} it in advance. This
is live terrain: ES is being actively re-proposed for LLM fine-tuning
\citep{sarkar2025hyperscale} and its failure modes actively studied
\citep{abdi2026esforgetting, hoy2026esgeometry}---which is precisely why a pre-build
gate is useful, since it tells you, before committing to an ES outer loop, whether a
cheap baseline already recovers the frontier.

\paragraph{Importance-based mask selection / pruning.} Magnitude pruning
\citep{han2015learning}, iterative magnitude pruning with rewind
\citep{frankle2019lottery, frankle2020linear}, connection sensitivity
\citep{lee2019snip}, and Fisher-information importance
\citep{kirkpatrick2017overcoming, sung2021training} produce strong masks from
single-shot or cheap iterative statistics. \citet{su2020sanity} go further: randomly
reshuffling \emph{which} weights are kept within each layer leaves final accuracy
unchanged, so only the per-layer sparsity ratios carry information and the mask
identity does not---the direct analog of our random-$\kappa$ control. That these rival
searched masks is established; our mask-identity result (Section~\ref{sec:maskid}) is a
confirmation in a bilevel-constrained-fitness setting, not a new claim.

\paragraph{Parameter-space constraints for ES on LLMs (concurrent).}
\citet{schweighofer2026overcoming} study parameter-space constraints (magnitude
anchoring) for ES on LLM fine-tuning---concurrent work in the same family, orthogonal
to mask-identity selection. Our mask-identity baseline includes the \textbf{generic
anchored-weight-decay mechanism}---a plain $L_2$ pull of the adapter toward its warm
checkpoint ($\gamma \sum \|W - W_\text{warm}\|^2$), which we ran as a baseline, not a
reimplementation of their specific formulation; we used their reported range only as
a starting estimate for the single strength hyperparameter $\gamma$. This generic
anchored weight decay ties the best single-shot mask method within noise---a
positioning point, not a contest.

\paragraph{``Is the search worth it?'' decision procedures.} The closest formal
relatives decide \emph{whether} an expensive search is worthwhile rather than how to
run it. \citet{weerts2020tuning} formalize a non-inferiority test and a ``tuning
risk''---the performance lost by leaving a hyperparameter at its default instead of
tuning it---and find defaults are often non-inferior; \citet{bahmani2021totune}
likewise recommend which hyperparameters are worth tuning. In neural architecture
search, random search is a competitive baseline \citep{li2019randomnas}, the search
phase often fails to beat random selection \citep{yu2019evalnas}, and
relative-improvement-over-random has been proposed as the only honest NAS metric
\citep{yang2020nashard}. The decisive difference is \emph{when} the cheap baseline is
consulted: in all of this work it is a \emph{post-hoc} comparator used to ground or
debunk a method after it is built. Ours is a \emph{pre-registered, pre-build}
go/no-go gate, with a committed recovery-ratio threshold, that decides whether to
construct the outer loop at all. The closest pre-build intuition we found---running
random search first to gauge a search space's potential \citep{krishna2021adsnas}---is
stated as informal advice, with no recovery ratio, threshold, or pre-registration. We
do not claim the underlying intuition is new; we claim its packaging as a falsifiable,
pre-committed decision rule is.

The screening rule itself---a pre-registered, metric-matched, pre-build decision
procedure with a falsifiable threshold---is, to our knowledge, not packaged in this
literature.

\section{The Screening Gate and Rule}
\label{sec:rule}

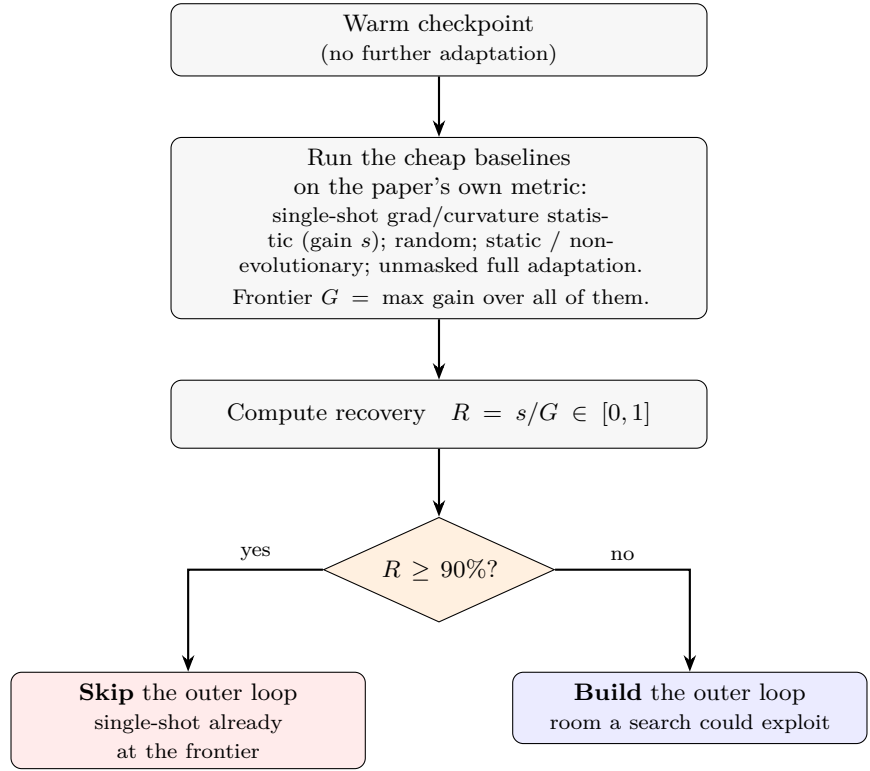
\begin{figure}[t]
\centering
\begin{tikzpicture}[
  node distance=8mm and 14mm,
  every node/.style={font=\small},
  box/.style={rectangle, rounded corners, draw=black, fill=black!3,
              text width=68mm, align=center, inner sep=4pt, minimum height=9mm},
  dec/.style={diamond, aspect=2.2, draw=black, fill=orange!12,
              text width=22mm, align=center, inner sep=1pt},
  res/.style={rectangle, rounded corners, draw=black, text width=44mm,
              align=center, inner sep=4pt, minimum height=9mm},
  arr/.style={-{Stealth[length=2.2mm]}, thick}
]
\node[box] (warm) {Warm checkpoint\\ \footnotesize(no further adaptation)};
\node[box, below=of warm] (base) {Run the cheap baselines on the paper's own
  metric:\\[1pt] \footnotesize single-shot grad/curvature statistic (gain $s$);
  random; static / non-evolutionary; unmasked full adaptation.\\[1pt]
  Frontier $G=\max$ gain over all of them.};
\node[box, below=of base] (R) {Compute recovery\quad $R = s / G \in [0,1]$};
\node[dec, below=9mm of R] (q) {$R \geq 90\%$?};
\node[res, fill=red!8, below left=10mm and 2mm of q] (skip)
  {\textbf{Skip} the outer loop\\ \footnotesize single-shot already at the frontier};
\node[res, fill=blue!8, below right=10mm and 2mm of q] (build)
  {\textbf{Build} the outer loop\\ \footnotesize room a search could exploit};
\draw[arr] (warm) -- (base);
\draw[arr] (base) -- (R);
\draw[arr] (R) -- (q);
\draw[arr] (q) -| (skip) node[pos=0.25, above, font=\footnotesize] {yes};
\draw[arr] (q) -| (build) node[pos=0.25, above, font=\footnotesize] {no};
\end{tikzpicture}
\caption{The Phase-0 Cheap-Baseline-Falsification gate. All gains are measured on
the paper's own metric over the warm (no-further-adaptation) baseline. The recovery
$R=s/G$ compares the best single-shot gradient/curvature statistic's gain $s$ to the
achievable frontier $G$ (the best gain reached by \emph{any} cheap method, single-shot
included). Defining $G$ as that maximum keeps $R$ a proper fraction and makes the
kill signal ``single-shot \emph{is} the best cheap method.''}
\label{fig:gate}
\end{figure}

\paragraph{Gate.} Before any outer-loop implementation, enumerate the cheapest
plausible non-mechanism explanations of the expected effect (random, single-shot
heuristic, static/non-evolutionary) and run them on the \emph{same metric the paper
will use}. The outer loop is built only if those baselines fail to capture the
effect. Figure~\ref{fig:gate} summarizes the procedure.

\paragraph{Screening rule.} All gains are measured on the paper's metric, over the
warm (no-further-adaptation) baseline. Let the \emph{achievable frontier} $G$ be the
best gain reached by any method evaluated at the Phase-0 gate (single-shot
statistics, random and static baselines, unmasked full adaptation), and $s$ the gain
of the best single-shot gradient/curvature statistic. Define
$R = s / G \in [0, 1]$. \textbf{If $R \geq 90\%$, skip the outer loop}---the
single-shot statistic already sits at (or defines) the frontier. Defining $G$ as the
max over all evaluated methods keeps $R$ a proper fraction and makes the kill signal
``single-shot \emph{is} the best cheap method.''

\begin{figure}[t]
\centering
\includegraphics[width=0.92\linewidth]{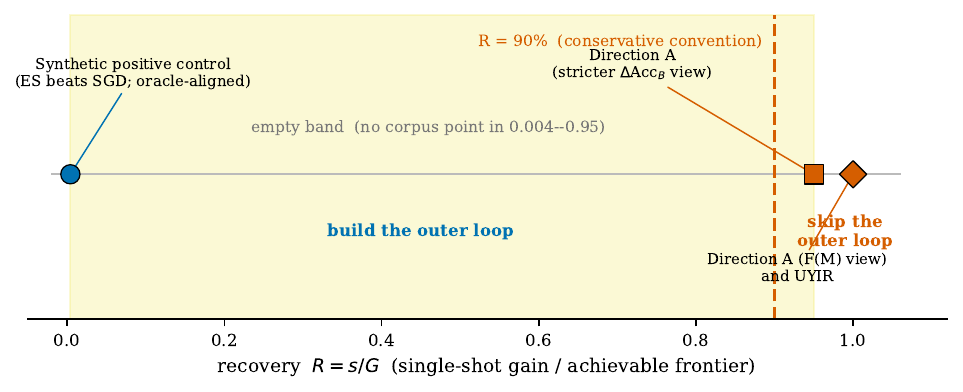}
\caption{The recovery axis and the conservative $90\%$ convention. The only corpus
case where an outer loop beat single-shot search sits at $R \approx 0.004$ (a
synthetic, oracle-aligned positive control; Section~\ref{sec:positive}); both real
kills sit at $R \approx 1.0$ (the mask-identity study also reads ${\approx}\,0.95$ under a
stricter $\Delta\text{Acc}_B$ view---a second view of the same kill). No point falls
in the empty band $(0.004, 0.95)$, so the data fixes a cutoff only somewhere inside
it; $90\%$ is chosen conservatively, not measured.}
\label{fig:raxis}
\end{figure}

\paragraph{Threshold anchoring.} Both corpus kills sit at $R \approx 1.0$
(the mask-identity study, Section~\ref{sec:maskid}, also reads ${\approx}\,0.95$ under a stricter
$\Delta\text{Acc}_B$ view---two views of \emph{one} case, not a spread across cases);
the only case where an outer loop beat single-shot had $R \approx 0.004$. The data
therefore supports \emph{any} cutoff in the wide empty band between ${\approx}\,0.004$
and ${\approx}\,0.95$; it does \textbf{not} locate $90\%$ precisely
(Figure~\ref{fig:raxis}). We adopt \textbf{$90\%$ as a deliberately conservative
convention}---when in doubt, prefer wrongly \emph{building} the outer loop over
wrongly \emph{skipping} it---and note that no corpus point falls in the ambiguous
band, so the convention affects no decision reported here. The rule is
\textbf{prospectively falsifiable}: a task with $R < 90\%$ where the outer loop still
fails to beat single-shot would refute it. Table~\ref{tab:corpus} collects the full
corpus the rule is validated against.

\begin{table}[t]
\centering
\caption{The within-lab corpus at a glance. Every real-task case screens
$R \geq 90\%$ and the outer loop is skipped or never built; the only $R < 90\%$
instance is an oracle-aligned synthetic control. $R$ is the recovery $s/G$
(Section~\ref{sec:rule}) unless noted.}
\label{tab:corpus}
\footnotesize
\begin{tabular}{@{}llllccl@{}}
\toprule
Case & Objective class & Metric & Cheap method at frontier & $R$ & $n$ & Decision \\
\midrule
Case 1 (MoE/LOO) & smooth MoE routing & held-out NLL & static router-substrate & ${\approx}1.0$ & 3 & skip \\
\S\ref{sec:kappa} $\kappa$-sparsity & smooth LoRA rescue & preservation & random-$\kappa$ control & ${\gtrsim}1.0^{\dagger}$ & 2 & skip \\
\S\ref{sec:maskid} mask identity & smooth bilevel $F(M)$ & $F(M)$ & gradient-norm single-shot & ${\approx}1.0\,(.95)$ & 3 & not built \\
Direction C (ICL) & discrete black-box & held-out acc. & greedy/beam construction & $0.97^{\ddagger}$ & 1 & not built \\
synthetic control & discrete, oracle-aligned & routing gap & \emph{none}: ES wins & $0.004^{\S}$ & --- & built \\
\bottomrule
\end{tabular}

\vspace{3pt}
{\footnotesize $^{\dagger}$A direct cheap-control-beats-outer-loop kill (control
$0.906 >$ rescue $0.805$), not an $s/G$ ratio.\quad
$^{\ddagger}$$R_{\text{screen}}$, the domain-appropriate result-ratio for a
no-gradient discrete objective (Section~\ref{sec:positive}).\quad
$^{\S}$The lone $R < 90\%$ instance where the outer loop wins---but oracle-aligned,
not a real task.}
\end{table}

\section{Case 1 --- A Separately Reported MoE-Lifecycle Project: LOO-Aligned Fitness (Cited)}
\label{sec:caseone}

This case \citep{priormoe2026} is a separate companion preprint, publicly available
on arXiv; we cite its reported results and do not reopen them.
Pre-registered claim: a
leave-one-out-aligned fitness signal in an evolutionary MoE lifecycle improves
outer-loop selection over a static mixture-of-experts.

On real text, every clean active-turnover comparison was a loss or a within-noise
tie against the static-MoE baseline. The one comparison reaching significance (code
domain, $+0.0625$ nats, $p=0.009$) is decomposed by that project's own factorial: the gain
\textbf{localizes to a static router-substrate change} ($+0.0976$ nats,
$p \leq 0.002$), while the evolutionary lifecycle term contributes \textbf{no
detectable gain} ($-0.028$ nats, $p \approx 0.05$ at $n=3$). The eval used
deterministic batching, which narrows the variance estimate and therefore inflates
the apparent significance of every term; the true uncertainty is wider than reported,
making ``no detectable gain'' the \textbf{conservative} reading (and the
router-rewrite significance an upper bound). We make no ``the lifecycle harmed
performance'' claim---the term is indistinguishable from zero. In screening terms: a
static (single-shot-class) computation was the frontier; $R \approx 1.0$; the gate
fires.

\section{Case 2 --- The Mask-Rescue Arc: Heritability \texorpdfstring{$\rightarrow$}{to} \texorpdfstring{$\kappa$}{kappa}-Sparsity \texorpdfstring{$\rightarrow$}{to} Mask Identity (This Paper)}
\label{sec:casetwo}

A single project pre-registered two successive outer-loop bets; both were falsified
by a cheap baseline on the project's own metric.

\subsection{Heritability does not rescue (the \texorpdfstring{$\kappa$}{kappa}-sparsity result)}
\label{sec:kappa}

Pre-registered claim: \emph{heritable} sparse mutability masks, evolved under
selection, rescue gradient-warmed LoRA adapters from perturbation better than
non-heritable controls. The decisive comparison is against a
\textbf{lifecycle-free random-$\kappa$ control} (fresh random $\kappa$-sparse masks
each epoch; no inheritance, no selection) and a dense matched-uniform baseline.

\paragraph{Provenance of the destructive regime.} The $\sigma=1\text{e-}1$ scale used
here was \emph{not} validated as a reproduction of the prior warm-start failure
boundary: a pre-registered matched-$\sigma$ replication (one-shot and faithful-ES
variants) failed its acceptance band at every sampled scale, and these rescue
comparisons proceed under a scoped decision to test rescue at a single
confirmed-destructive scale rather than across a replicated boundary. This does not
affect the screening conclusion---the cheap random-$\kappa$ control is compared
against the outer loop at the same scale---but it bounds the external reading: this is
rescue at a fixed destructive scale, not across a reproduced failure curve.

\begin{figure}[t]
\centering
\includegraphics[width=\linewidth]{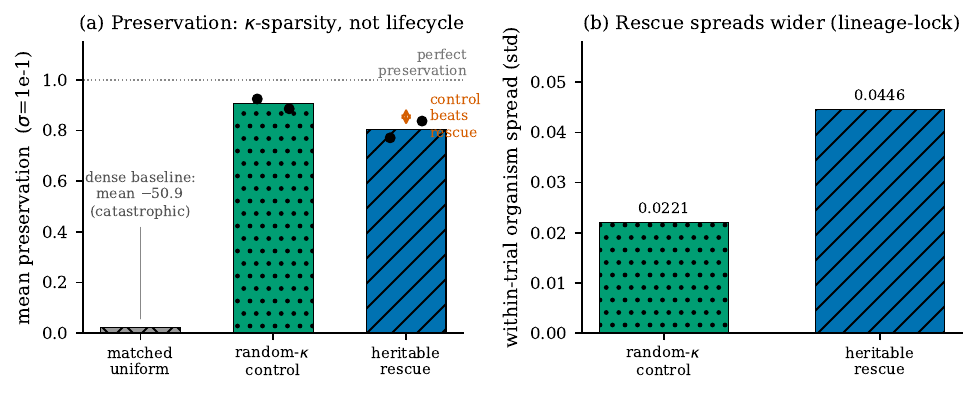}
\caption{The mask-rescue project at the only discriminating perturbation scale
($\sigma=1\text{e-}1$; $n=2$ seeds, dots are per-seed values). \textbf{(a)} Mean
preservation: the lifecycle-free random-$\kappa$ control (0.906) \emph{beats} the
heritable rescue loop (0.805); the dense matched-uniform baseline is catastrophic
(off scale). The rescue effect is $\kappa$-sparsity, not heritability or selection.
\textbf{(b)} Within-trial organism spread: heritable rescue spreads \emph{wider}
than the control, consistent with the lifecycle locking early-unlucky lineages. At
$\sigma \leq 1\text{e-}2$ both sparse protocols preserve $\approx 1.0$ and are
indistinguishable.}
\label{fig:rescue}
\end{figure}

\begin{figure}[t]
\centering
\includegraphics[width=0.92\linewidth]{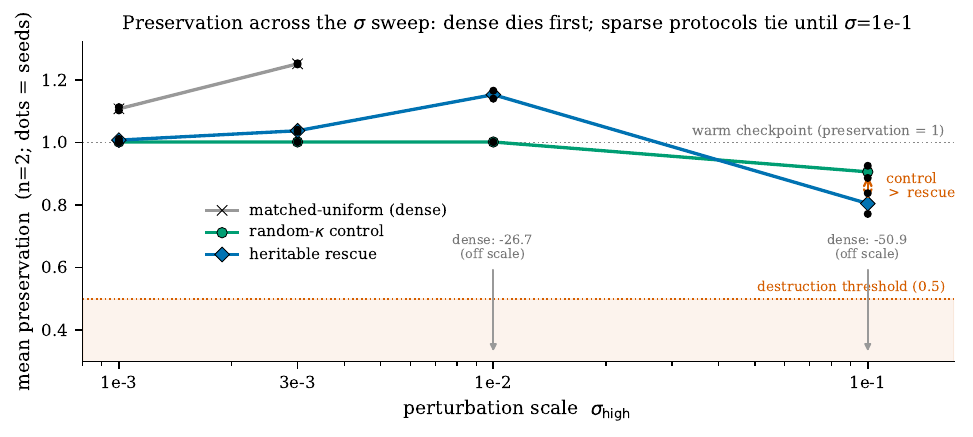}
\caption{Preservation across the full $\sigma_{\text{high}}$ sweep, all three
protocols, clean reruns ($n=2$ seeds 42/43; black dots are per-seed values). The
dense matched-uniform baseline preserves at $\sigma \leq 3\text{e-}3$ but is
catastrophically destroyed at $\sigma \geq 1\text{e-}2$ (off scale; values
annotated). Both sparse protocols hold near the warm checkpoint through
$\sigma=1\text{e-}2$ and diverge only at $\sigma=1\text{e-}1$, where the
lifecycle-free random-$\kappa$ control overtakes the heritable rescue---the single
discriminating scale of Figure~\ref{fig:rescue} is the rightmost point here. The
result substantiates in one view the regime claims made in prose: dense dies first,
sparse ties below the discriminating scale, control $>$ rescue only at the top.}
\label{fig:sigmasweep}
\end{figure}

At the only perturbation scale that discriminates ($\sigma=1\text{e-}1$), the
random-$\kappa$ control \textbf{beat} the heritable outer loop on preservation
(\textbf{0.906 vs 0.805}, sign-consistent across both available seeds), and the
heritable loop also showed wider within-trial spread (Figure~\ref{fig:rescue}). At
$\sigma \leq 1\text{e-}2$ the two sparse protocols tie (perturbation too small to
test the mask); in the safe regime the dense baseline preserves \emph{more} than the
heritable loop. The heritable loop's only ``wins'' are against the dense strawman
($+52$ pp at $\sigma=1\text{e-}1$) and are fully attributable to
\textbf{$\kappa$-sparsity}, which the lifecycle-free control reproduces. Heritability
and selection---the evolutionary content---add nothing; the gate fires on the
random-$\kappa$ control. These cross-seed results are $n=2$ (seeds 42, 43); we report
the ordering as sign-consistent and treat the magnitudes as provisional. The full
$\sigma$ sweep (Figure~\ref{fig:sigmasweep}) shows the regime structure directly: the
dense baseline collapses first---catastrophic at $\sigma \geq 1\text{e-}2$---while both
sparse protocols hold near the warm checkpoint through $\sigma=1\text{e-}2$ and
separate only at $\sigma=1\text{e-}1$, the single scale where the control overtakes the
rescue.

\subsection{Mask identity does not carry exploitable information (the mask-identity study)}
\label{sec:maskid}

Pre-registered claim: the \emph{identity} of the optimal $\kappa$-sparse mask carries
causal information an evolutionary search can exploit, beyond what single-shot
statistics recover, under a bilevel constrained fitness $F(M)$.

\begin{figure}[t]
\centering
\includegraphics[width=\linewidth]{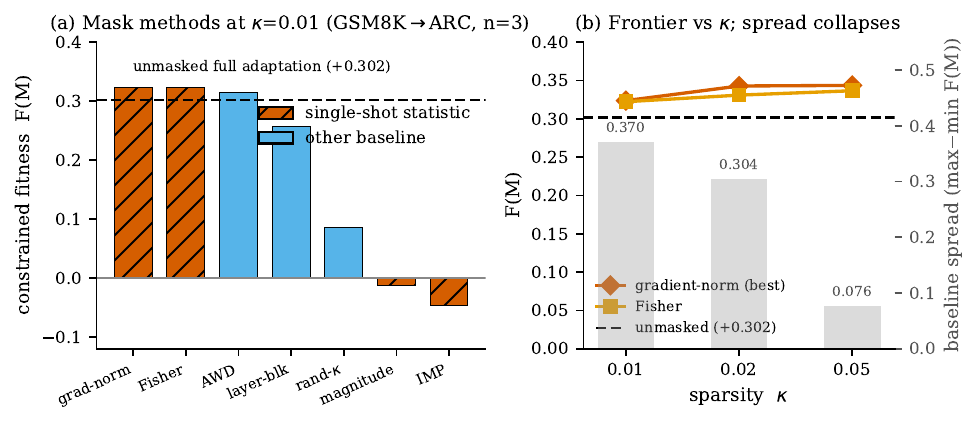}
\caption{Mask-identity baseline-kill on GSM8K$\rightarrow$ARC-Challenge.
\textbf{(a)} Per-protocol mean constrained fitness $F(M)$ at $\kappa=0.01$ ($n=3$).
The two single-shot statistics (gradient-norm, Fisher; hatched) sit \emph{above} the
unmasked full-adaptation frontier ($+0.302$, dashed); magnitude and IMP actively
hurt. \textbf{(b)} The single-shot frontier holds across $\kappa$ (gradient-norm and
Fisher both exceed $+0.302$ at every $\kappa$), while the baseline spread---the
signal a mask search could exploit---collapses from $0.370$ to $0.076$ as $\kappa$
grows. There is no headroom for an evolved mask at any sparsity.}
\label{fig:maskid}
\end{figure}

On GSM8K$\rightarrow$ARC-Challenge at $\kappa \in \{0.01, 0.02, 0.05\}$, the best
single-shot statistic mask (gradient-norm, $F(M) = +0.324 / +0.343 / +0.344$) was
\emph{itself the frontier}---it exceeded even unmasked full adaptation ($+0.302$) at
every sparsity (\textbf{$R \approx 1.0$}; ${\sim}\,0.95$ on the stricter
$\Delta\text{Acc}_B$ view), leaving ${\approx}\,0$ headroom (Figure~\ref{fig:maskid}).
Baseline spread collapses with $\kappa$ ($0.370 \rightarrow 0.304 \rightarrow 0.076$):
mask identity matters \emph{less} with more capacity, so $\kappa=0.01$ is the hardest
test and showed no headroom even there. Under the screening rule, the evolved-mask
outer loop (Protocol~7) was \textbf{not built}.

\paragraph{Evidence bar.} The kill was declared on a single task pair
(ARC-Challenge; the default Far pair, HumanEval, failed the pre-flight tractability
gate and was substituted---disclosed). The pre-registered rule requires a
\emph{positive} result to hold across all three pairs (no cherry-picking); a
\emph{negative} is conclusive from any pair where the single-shot baseline already
exceeds the ceiling, since additional pairs cannot create headroom the easiest cell
lacks.

\section{What the Gate Bought}
\label{sec:ledger}

For the mask-identity study (Section~\ref{sec:maskid}), the Phase-0 gate (pre-flight + budget
calibration + baseline-kill across three $\kappa$) cost ${\sim}50$--$70$ GPU-hours on
a single consumer GPU and returned a $\kappa$-robust kill. Building and running the
screened-out evolutionary outer loop was estimated at ${\sim}400{+}$
GPU-hours\footnote{Estimated as population ($\sim$16) $\times$ generations
($\sim$50) $\times$ per-organism adapt-and-evaluate cost ($\sim$0.5 GPU-hour, the
measured per-trial cost of the baseline-kill runs on the same hardware) $\approx 400$
GPU-hours. This is an order-of-magnitude estimate of work deliberately never run, not
a measurement.} for the first cell alone, before any multi-pair / $\kappa$-robustness
sweep, plus weeks of
implementation---a ${\sim}6$--$8\times$ GPU saving on the first cell, far more once
the unbuilt sweep and engineering time are counted (Table~\ref{tab:ledger}). The
recurring shape across all cases: a static or single-shot computation captured the
effect on the project's own metric, and it was identifiable \emph{before} the outer
loop was built.

\begin{table}[t]
\centering
\caption{What the Phase-0 gate bought on the mask-identity study. Costs are GPU-hours on a single
consumer GPU; the screened-out figure is the first-cell estimate only and excludes
the multi-pair / $\kappa$-robustness sweep and weeks of implementation that the kill
also foreclosed.}
\label{tab:ledger}
\begin{tabular}{lr}
\toprule
Item & GPU-hours \\
\midrule
Phase-0 gate (pre-flight + budget calibration + baseline-kill $\times 3\,\kappa$) & ${\sim}50$--$70$ \\
Screened-out evolutionary outer loop (first cell only) & ${\sim}400{+}$ \\
\midrule
First-cell saving & ${\sim}6$--$8\times$ \\
\bottomrule
\end{tabular}
\end{table}

\section{The Positive Condition --- a Hypothesis, Not a Result}
\label{sec:positive}

We state the organizing thesis explicitly as a \textbf{post-hoc synthesis}, not a
pre-registered prediction: an evolutionary outer loop earns its keep only when it
optimizes structure the inner optimizer cannot reach---a non-differentiable, discrete,
or black-box objective gradient descent does not serve and no single-shot statistic
recovers ($R < 90\%$ at the gate).

\begin{figure}[t]
\centering
\includegraphics[width=0.56\linewidth]{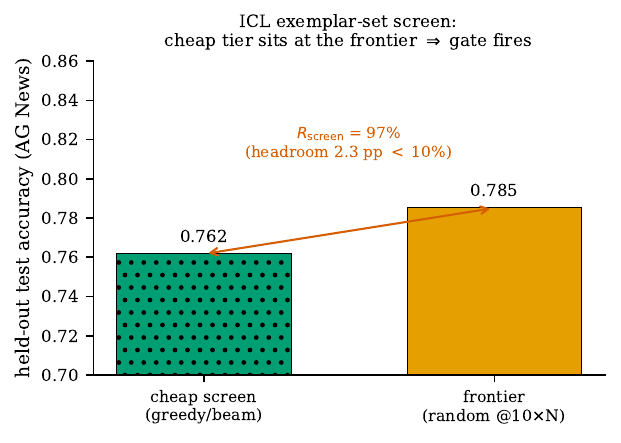}
\caption{The Direction C in-context-learning exemplar-set screen (AG News,
Qwen2.5-1.5B-Instruct, single seed). The cheap greedy/beam-with-restarts
construction (0.762 held-out test accuracy) recovers $R_{\text{screen}} \approx 97\%$
of the frontier reached by random search at the outer loop's own matched budget
($10\,N_{\text{screen}}$; 0.785). The $2.3$-pp headroom is far below the $10\%$ the
rule requires, so the gate fires and the evolutionary outer loop was not built. This
is the first deliberate hunt for an $R<90\%$ real-task substrate, and its first
candidate screened $R \geq 90\%$.}
\label{fig:iclscreen}
\end{figure}

We have \textbf{zero real-task instances} of this condition in the present corpus. The
only positive evidence is a synthetic mechanism existence-proof.\footnote{A prior
synthetic sandbox in which ES beat SGD on a discrete router by ${>}3\sigma$ (closing
$55.9\%$ of the routing gap vs SGD's $0.2\%$)---but only because the
adapter$\rightarrow$domain mapping was oracle-pre-aligned; removing the oracle
collapses ES to the no-skill floor. Its authors label it a positive control, not a
co-evolution result.} It demonstrates the mechanism under an idealized condition; it
is not a task win. A \textbf{pre-registered successor design} was built to \emph{hunt}
for exactly this: a real, discrete/non-differentiable objective that screens
$R < 90\%$ and on which an outer loop then wins. Its first candidate substrate has now
been screened---in-context-learning exemplar-set selection (choosing and ordering an
8-shot set from a 64-candidate pool to maximize held-out accuracy; AG News,
Qwen2.5-1.5B-Instruct), the discrete black-box regime this section names. The Phase-0
screen returned an analogous recovery $R_{\text{screen}} \approx 97\%$: the strongest cheap-tier method
(beam/greedy-with-restarts marginal-gain construction) recovered $97\%$ of the frontier
reached by random search at the outer loop's own matched budget ($10\,N_{\text{screen}}$),
so the $R_{\text{screen}} \ge 90\%$ gate fired and \textbf{the outer loop was not built}
(Figure~\ref{fig:iclscreen}).\footnote{Here
$R_{\text{screen}}$ is the pre-registered, domain-appropriate \emph{result}-ratio (best swept-cheap
result)/(achievable frontier) for a discrete objective with no gradient statistic---not
the gain-ratio $R = s/G$ of Sections~\ref{sec:rule}--\ref{sec:casetwo}, though both express that a cheap
method already sits at the frontier. Single seed; the screen sits ${\approx}7$ pp above
the gate.} This direction is independently documented---cheap retrieval's advantage
over a naive exemplar set collapses as the shot count grows \citep{bertsch2024longicl},
while set-level optimization helps mainly on tasks where cheap baselines leave headroom
\citep{ye2023ceil}, the very kind of task this hunt was designed to find. This is the
first \emph{deliberate hunt} for an $R<90\%$ real-task substrate,
and the first candidate came up $R_{\text{screen}} \ge 90\%$---a consistent confirmation that
\textbf{strengthens} the ``zero real-task instances'' finding above, not a refutation
(which would require $R<90\%$ with the outer loop still failing). A second
pre-registered branch (a circuit-discovery objective) is unrun. We stress that engaging
this condition is itself a substantial research effort---a full discrete-objective
evolutionary study with its own multi-week build and compute budget, deferred to future
work---not a cheap confirmation withheld. The paper stands entirely on the screening
rule and the corpus in hand; the successor study is invoked only as the falsifiable test
the rule invites, now with its first screen reported.

\section{Limitations}
\label{sec:limitations}

\begin{itemize}
\item \textbf{Not independent.} Same lab, author, and design instinct; the two cases
  are not independent replications. We present a series, and disclose \textbf{four
  further evolutionary efforts that produced no analyzable comparison}---an honest
  file drawer---plus \textbf{one further evolutionary outer loop that did produce an
  analyzable negative} and is not developed as a case: a faithful evolution-strategy
  boundary replication that failed to reproduce the assumed warm-start destruction
  boundary within its pre-registered band (it over- or under-shot at every sampled
  scale). It is consistent with the thesis---another outer loop that did not beat its
  cheap reference---and we keep it out of the corpus count only to avoid conflating a
  replication failure with the screening-rule cases (its destructive-regime
  consequence is noted in Section~\ref{sec:kappa}).
\item \textbf{Post-hoc thesis.} Each project pre-registered a \emph{narrower}
  mechanism (fitness alignment / heritability / mask identity), not the reachability
  law; the law is an after-the-fact synthesis, offered as a falsifiable hypothesis
  (Section~\ref{sec:positive}).
\item \textbf{Statistical strength.} Cross-seed results are $n=2$--$3$; we cite only
  out-of-noise effects as confirmations and label within-noise differences as ties.
\item \textbf{Novelty boundary.} The per-result findings echo known results
  (Section~\ref{sec:related}). The novel claim is the screening rule and its
  validation as a pre-build decision procedure.
\item \textbf{Single-substrate screening.} The rule is validated on LoRA/MoE-scale
  LLM work; generalization to other model families/scales is untested.
\end{itemize}

\section{Discussion and Conclusion}

The screening rule reframes a recurring, expensive mistake---building an evolutionary
outer loop over a space the inner optimizer already serves---as a \emph{cheap,
advance, falsifiable check}. Its value is decision-theoretic: a few GPU-days, run
before implementation, that can foreclose a multi-week build. Across a within-lab
series, both analyzed cases fired the gate, and a companion factorial
(Case~1) localizes the failure to the outer loop specifically rather than to the
substrate it rode on.

We deliberately decline the stronger story. There is no real-task positive in this
corpus; the positive condition is a hypothesis with a named, pre-registered
successor design, whose first screen is reported here and whose circuit branch
remains unrun. What we claim is narrower and, we argue, more useful: a
prospectively-falsifiable rule for \emph{when not to spend} on an evolutionary outer
loop, validated where it cost us the most to learn it.

\bibliographystyle{tmlr}
\bibliography{references}

\end{document}